\documentclass{article}

\usepackage{arxiv}

\usepackage[utf8]{inputenc} 
\usepackage[T1]{fontenc}    
\usepackage{hyperref}       
\usepackage{url}            
\usepackage{booktabs}       
\usepackage{amsfonts}       
\usepackage{nicefrac}       
\usepackage{microtype}      
\usepackage{lipsum}
\usepackage[english]{babel}
\usepackage{graphicx}
\usepackage{float}

\newcommand\Tstrut{\rule{0pt}{2.0ex}}         

\title{Optimal Use of Experience in First Person Shooter Environments}

\author{
Matthew Aitchison\\
College of Engineering and Computer Science\\
Australian National University\\
Canberra, Australia\\
\texttt{Matthew.Aitchison@anu.edu.au}
}

\newcommand{\note}[1]{\textbf{[*]}}

\usepackage[style=ieee,backend=biber]{biblatex}
\begin{document}

\begin{minipage}{\textwidth}\ \\[12pt]
978-1-7281-1884-0/19/\$31.00 \copyright 2019 IEEE
\end{minipage}

\maketitle

\begin{abstract}
Although reinforcement learning has made great strides recently, a continuing limitation is that it requires an extremely high number of interactions with the environment. In this paper, we explore the effectiveness of reusing experience from the experience replay buffer in the Deep Q-Learning algorithm. We test the effectiveness of applying learning update steps multiple times per environmental step in the VizDoom environment and show first, this requires a change in the learning rate, and second that it does not improve the performance of the agent. Furthermore, we show that updating less frequently is effective up to a ratio of 4:1, after which performance degrades significantly.  These results quantitatively confirm the widespread practice of performing learning updates every 4th environmental step.
\end{abstract}
\par

\section{Introduction}

Reinforcement learning (RL) in games has gained a lot of attention recently due to some high-profile successes such as super-human performance in the Atari games \cite{Mnih2015Human-levelLearning}, StarCraft \cite{Ontanon2013AStarCraft}, and the game of Go \cite{Silver2018ASelf-play.}. However, these algorithms require an extraordinary amount of experience to train the agents. AlphaZero, for example, played almost 5-million games to reach peak performance.  This level of sample inefficiency limits the application of these algorithms to real-world problems. In situations where generating experience is costly, for example when robotic interaction with the world is required, episodes typically measure in thousands rather than millions \cite{Gu2016DeepUpdates}. Interest has been building for sample efficient algorithms with the introduction of the MineRL environment \cite{Guss2019ThePriors}, as this is of critical importance to the application of RL to real-world robotic problems. 

The Deep Q-Learning Network (DQN) algorithm \cite{Mnih2015Human-levelLearning} first demonstrated that Deep Neural Networks (DNNs) could be successfully applied to complex RL tasks. To do this, thr algorithm implemented an Experience Replay (ER)  \cite{Lin1992Self-ImprovingTeaching} system that allowed the agent to sample randomly from previously experienced transitions. Although the primary purpose was to decorrelate the samples, a useful side effect is the ability to learn from experience multiple times. The original paper performed one learning step on 32 samples for every environmental step, meaning that each transition would be seen an average of 32 times. Later papers instead apply updates every 4th environment step giving transitions an expected sample rate of 8 times \cite{VanHasseltDeepQ-learning} \cite{Lample2016PlayingLearning}. 

\begin{figure}
    \centering
    \includegraphics[width=0.48\textwidth]{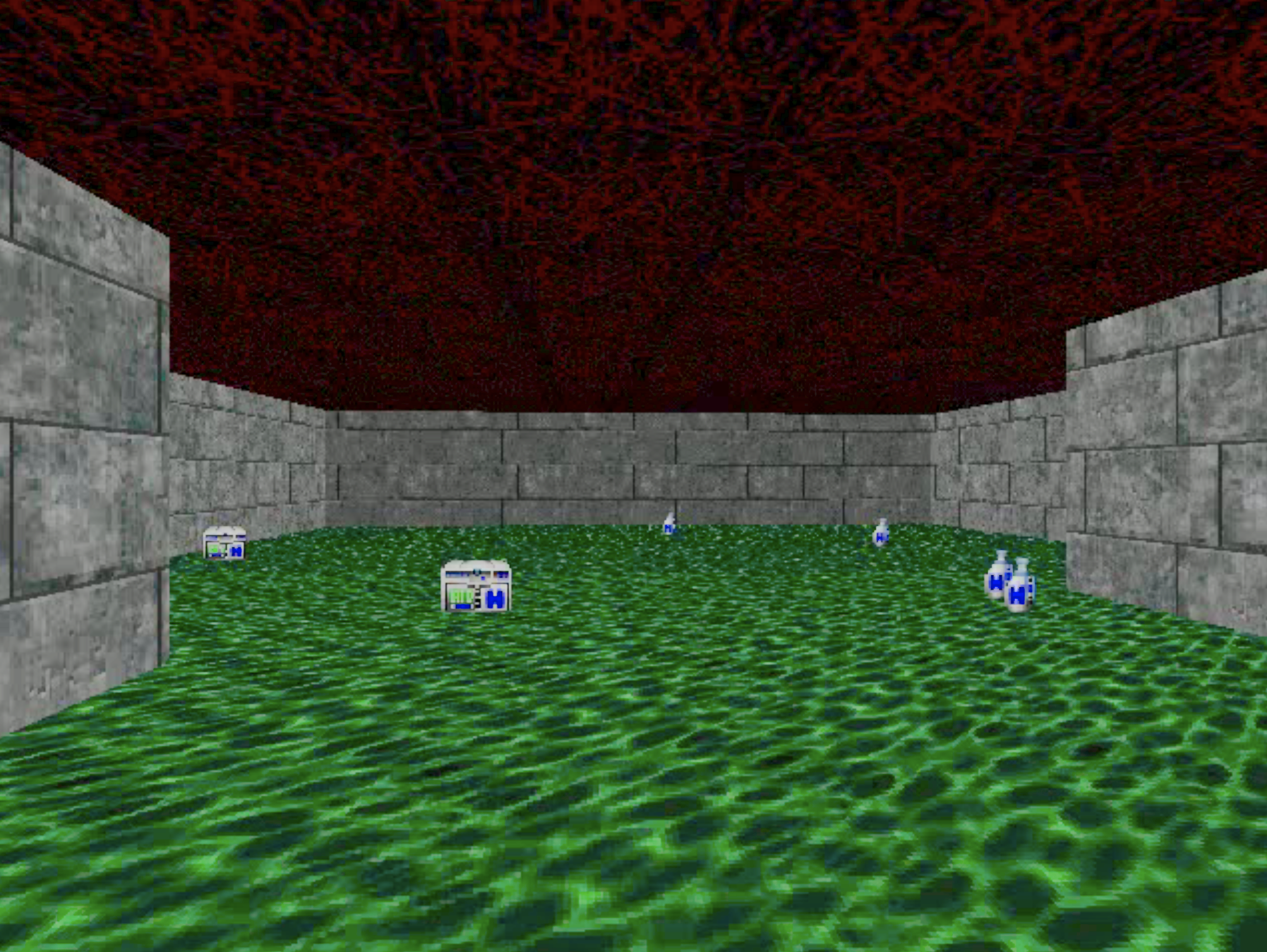}
    \caption{ViZDoom `Health Gathering Supreme' scenario showing health kits and poison bottles.}
    \label{fig:vizdoom}
\end{figure}

To our knowledge, this change has never been quantitatively assessed in a complex 3D environment. In this paper, we investigate the impact of adjusting the learning update ratio for DQN under a First Person Shooter 3D environment.  If increased sampling could make up for environmental steps training schemes may be able to apply additional model updates in lieu of environmental steps where the latter has a significantly higher cost.  Furthermore, in situations where the cost of generating experience is relatively low, a lower frequency may improve training times with little impact on the performance of the agent.

The `Health Gathering Supreme' scenario in the ViZDoom environment was used to evaluate agents performance under different experience replay sampling ratios \cite{Kempka2016ViZDoom:Learning}. The scenario tests navigating and fine-grained control in a challenging, rich, 3D environment as shown in figure \ref{fig:vizdoom}. Furthermore, the graphics of ViZDoom more closely represent a real-world setting than other environments such as the Atari Learning Environment (ALE) \cite{Bellemare2012TheAgents}.

\begin{figure*}[ht]
    \centering
    \includegraphics[width=0.95\textwidth]{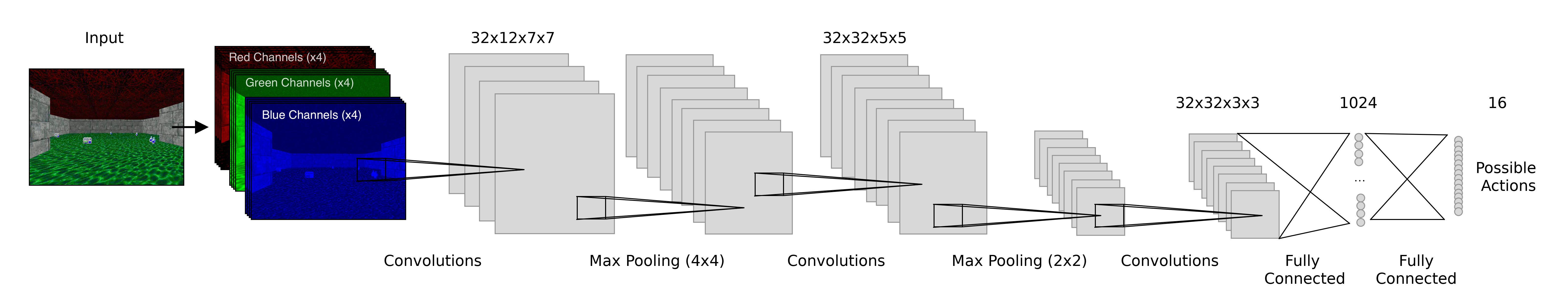}
    \caption{Architecture of the model used in these experiments.}
    \label{fig:model}
\end{figure*}

\section{Related Work}

Various sample efficient algorithms have been proposed to address the shortcomings of the DQN algorithm. These algorithms, however, would also benefit from knowing the optimal ratio between taking environmental steps and learning steps.

\subsection{A3C / A2C}
The Asynchronous Advantage Actor-Critic (A3C) algorithm and its synchronous version A2C were designed to improve the computational efficiency of the Actor-Critic algorithm by allowing agents to run in parallel \cite{Mnih2016AsynchronousLearning}.  This greatly improves training time but has poor sample efficiency due to the lack of an experience replay.

\subsection{Prioritized Experience Replay}
Prioritized experience replay reuses experiences from the past according to their importance \cite{Schaul2015PrioritizedReplay}. This importance is estimated from the temporal difference (TD) error, where larger errors are believed to be more important.  Importance sampling greatly increases the efficiency of an experience replay buffer, however the question of how many times to sample the replay buffer remains.

\subsection{ACER}
The Actor-Critic with Experience Replay (ACER) algorithm extends A2C to use an experience replay and shows that mixing off-policy experience replay updates with on-policy updates provides much better sample efficiency \cite{Wang2016SampleReplay}. This approach differs from ours in that is mixes off-policy updates with on-policy, while ours uses only off-policy updates.

\section{Method}

In this section, we outline the experimental method for evaluating the optimal learning step ratio in the ViZDoom `Health Gathering Supreme' environment. Where the learning step ratio is the number of times to apply a learning update per environmental step.  Ratios less than 1 indicate multiple environmental steps per learning update.

\begin{table}[]
    \centering
    \caption{Hyperparameter settings}
    \begin{tabular}{l r}
        \hline
        Hyperparameter & Value \\
        \hline
        Minibatch size & 32 \Tstrut\\
        Replay memory size & 10,000 \\
        Target network update frequency & 1,000 \\
        Discount factor & 1.0 \\
        Initial exploration rate & 1 \\
        Final exploration rate & 0.1 \\
        Frame skip & 10 \\
        \hline
        
    \end{tabular}
    \label{tab:hyperparams}
\end{table}

\begin{figure*}[h]
    \centering
    \includegraphics[width=0.9\textwidth]{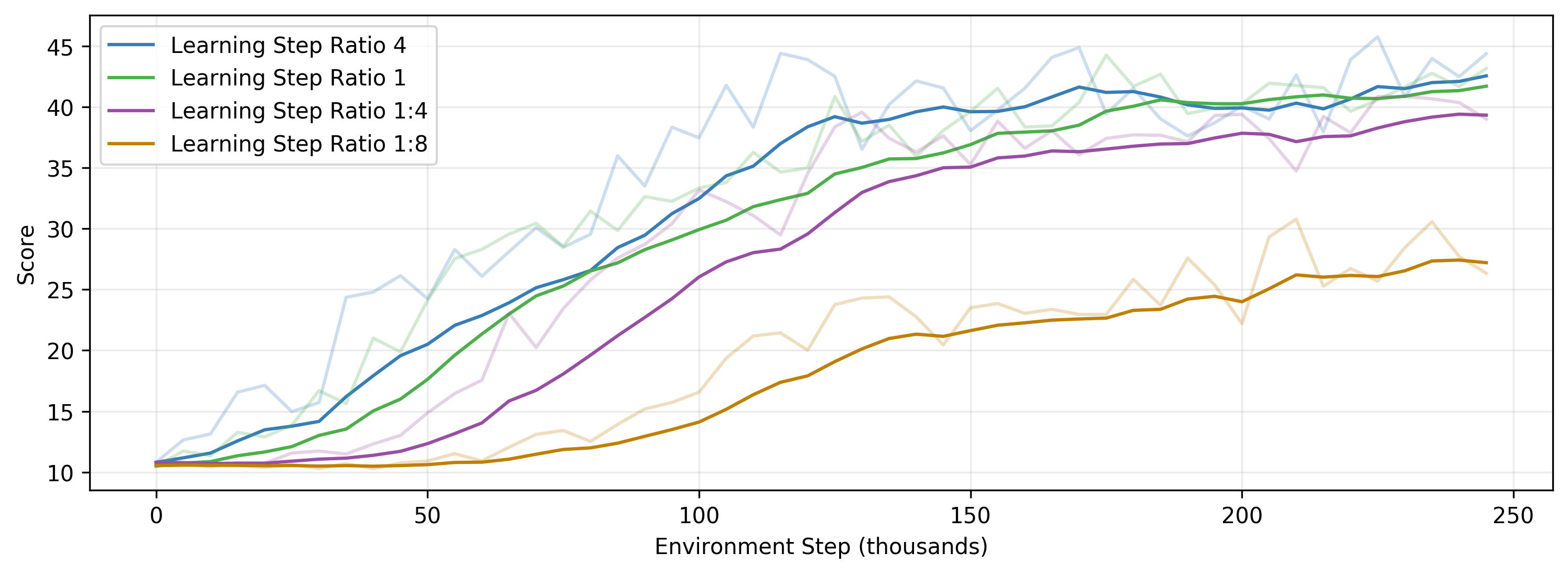}
    \caption{Agents test scores during training taken from the optimal learning rate, and averaged over the 5 runs. The every 4th step agent takes longer to start, but ends up only slightly underperforming the other agents even with significantly fewer learning updates. Raw scores are shown faded, with a smoothed with an exponential moving average ($\epsilon=0.8$) overlaid.}
    \label{fig:training}
\end{figure*}

\subsection{ViZDoom Environment}

The environment selected was `Health Gathering Supreme' which provides a substantial challenge to both AI agents and human players alike \cite{Kempka2016ViZDoom:Learning}. The scenario involves navigating a maze while collecting health kits and avoiding poison bottles (see figure \ref{fig:vizdoom}).  At intervals, the agent's health is decreased making it imperative to obtain health kits continuously. The scenario settings were left unchanged.\footnote{The VizDoom paper uses a backwards action which is not part of the default configuration for this environment, so one was added.  Also, the screenshots from the VizDoom paper appear to use a different texture pack as the poison bottles have a different colour.  The texture of the bottle in ViZDoom V1.1.7 is more difficult to distinguish from health than the reference screenshot in their paper}

\subsection{Reward Shaping}

We use change in health as the reward function as per \cite{Dosovitskiy2016LearningFuture}, as well as adding an auxiliary reward of +100 for each health pack and -100 for each poison bottle.  As health is capped at 100, this change slightly reduces the reward for obtaining health packs at high health levels.  The final score for the agent is calculated as the average health over the episode, with any time spent dead counted as 0.  This better distinguishes between an agent who maintains high health over the entire episode from one who only just survives.  Comparison is also made with Kempka et al's original agent according to the default scoring system. Using our scoring system an idle agent receives an average score of 11.6, whereas a theoretically perfect score would be close to 100.0.

\subsection{Network Architecture}

The network architecture used is nearly identical to the model used in \cite{Kempka2016ViZDoom:Learning} and is described in figure \ref{fig:model}. 

\subsection{Hyperparameters}

Hyperparameters were left largely unchanged from the ViZDoom paper \cite{Kempka2016ViZDoom:Learning}, with the following changes: A minibatch size of 32 was used to better match the DQN paper. Also, a target network updating scheme was implemented as per \cite{Mnih2015Human-levelLearning}, as this was found to improve training stability.  A full list of hyperparameters can be found in table \ref{tab:hyperparams}. For each learning step ratio, $\alpha$, a learning rate search was performed by selecting learning rates from 
\begin{equation}
\{5 \cdot 10^{-5} \cdot \alpha \cdot 2^{k} \hspace{0.5em} | \hspace{0.5em} k \in [-2, -1, 0, 1, 2]\}
\end{equation}
and choosing the learning rate with the highest final score.

Even though the VizDoom paper used 1,000,000 environment steps to train their model we found that by including target updates, most of our models converged much faster than this and did not improve much after the 150,000 mark, so trained for 250,000 environment steps.  The shorter environment steps helped with training times, and better represents a scenario in which generating environment steps is expensive.

\subsection{Evaluation}

Agents were evaluated by running the agent through the scenario 25 times every 5,000 environment steps, each time starting from a random position and orientation in the maze.  The scores over these episodes are then averaged. Due to fluctuations over time the results are then smoothed using an exponential moving average (EMA) with $\epsilon = 0.8$, and the top 10\% results averaged to give a final score. Results for each experiment were repeated 5 times with the average of these scores being presented.

\section{Experiments}

\subsection{Experimental Setup}

Experiments were performed with ViZDoom 1.1.7, on an Nvidia P100 GPU using PyTorch. We used the RMSProp optimizer with various learning rates and trained for 250,000 environmental steps.  Target model updates were performed every 1,000 learning steps. Training times ranged from 1 to 12 hours depending on the learning step ratio used.

\subsection{Experimental Results}

\begin{table}[ht]
    \centering
    \caption{Test results for agents with various learning step ratios.} 
    \begin{tabular}{c r c c}
        \hline
        Learning Ratio & Learning Rate & Score & Reward \\
        \hline
        4:1 & $1.25 \times 10^{-5}$ & 42 & 1142 \Tstrut\\ 
        2:1 & $2.5 \times 10^{-5}$ & 41 & 1132 \\
        \textbf{1:1} & $\mathbf{5 \times 10^{-5}}$  & \textbf{42} & \textbf{1163} \\
        1:2   & $2 \times 10^{-4}$   & 41 & 1129 \\
        1:4  & $4 \times 10^{-4}$   & 38 & 1015 \\
        1:8   & $2 \times 10^{-4}$ & 31 & 903 \\
        1:16   & $4 \times 10^{-4}$ & 24 & 699\\
        1:32   & $2 \times 10^{-4}$ & 18 & 538 \\
        \hline

    \end{tabular}
    \label{tab:results}
\end{table}

\begin{figure}[H]
    \centering
    \includegraphics[width=0.45\textwidth]{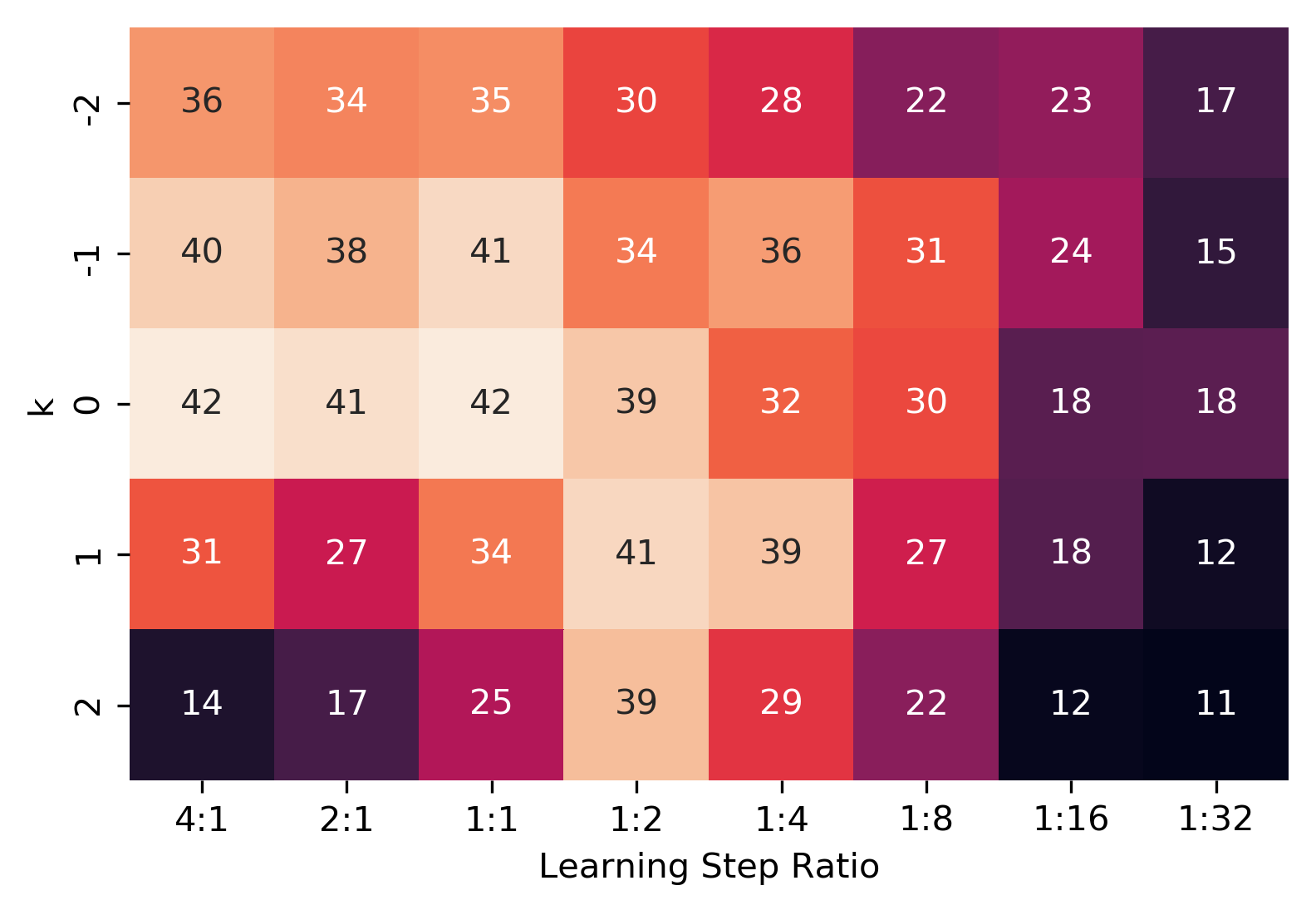}
    \caption{Heat map showing average final score for of the 5 runs over each learning rate modifier $k$ from eq. 1 and learning step ratio. A learning step ratio of 4:1 indicates 4 learning updates per environmental step, whereas a ratio of 1:4 indicates learning updates every 4th step.}
    \label{fig:heat_map}
\end{figure}

The agent's performance for each learning step ratio, is given in table \ref{tab:results}. Learning rate is the optimal learning rate tested, score the average final score over the 5 runs, and reward the average final reward over the 5 runs as per the ViZDoom paper. The agents performed similarly to those tested by Kempka et al, with the best run of the learning update every 4th environment step agent scoring 1,374 compared to approximately 1,300 in the original paper. The average results, however, are brought down by some runs failing to converge well. 

Figure \ref{fig:training} shows the average test performance during training for learning step ratios 4:1, 1:1, 1:4 and 1:8. Applying learning updates every 4th step does initially slow down training, however, this effect is minimal after 200,000 environment steps.  Applying learning updates every 8th step, on the other hand, significantly degrades performance.

Figure \ref{fig:heat_map} records the average score over the 5 runs for each learning rate modifier $k$ and learning step ratio. These results show that an increased learning step rate, by its self, is not sufficient to improve the performance of the model.  Indeed, simply applying learning updates additional times leads to a rapid decrease in performance. However, if the learning rate is similarly decreased by the equivalent amount, the agent's performance remains comparable. Notably, the learning step ratio can also be decreased to every 4th environment step with little impact on performance, again, so long as the learning rate is adjusted accordingly.
The reason for the agent’s sensitivity to the learning rate is not clear. One possible explanation could be that a certain amount of progress is needed per environment step, and it does not matter if this progress performed with multiple small steps or one large step.

\section{Conclusions and Future Work}

These results show that increased sampling of experience from the experience replay buffer does not improve an agent's performance, nor does it increase the speed at which an agent learns. On the other hand, decreasing the learning step ratio to a rate of every 4th environmental step has only a small negative effect on performance. Reductions beyond this point, however, severely degrade the agent's performance. When adjusting the learning step ratio it is important to also adjust the learning rate by an equal amount to maintain optimal performance.  These results quantitatively support for the `rule of thumb' of applying updates every 4th environmental step, with only a minimal impact on performance. These results suggest further research into the effect of learning update ratios on other algorithms that make use of Experience Replay, such as the ACER algorithm.

\section*{Acknowledgement}

This research was supported by an Australian Government Research Training Program (RTP) Scholarship.

\printbibliography

\end{document}